\pdfoutput=1

\documentclass[journal]{IEEEtran}

\ifCLASSINFOpdf

\else

\fi

\usepackage{cite}
\usepackage{lipsum}
\usepackage{tikz}
\usetikzlibrary{shapes,arrows}
\usepackage{standalone}
\usepackage{adjustbox}
\usepackage{amsmath,amssymb,amsfonts, mathtools}  
\usepackage{algorithmic}
\usepackage{graphicx}
\usepackage{textcomp}
\usepackage{xcolor}
\usepackage{float}
\usepackage{tabularx,booktabs}
\usepackage[font=small,labelfont=bf]{caption} 
\usepackage[subrefformat=parens]{subcaption}
\usepackage{hyperref}
\usepackage{pdfpages}
\usepackage{multirow,makecell}

\usepackage[roman]{parnotes}
\usepackage{algorithm2e,algorithmic}
\newcolumntype{C}{>{\centering\arraybackslash}X}

\begin{document}

\title{Adversarial Multi-Source Transfer Learning in Healthcare: Application to Glucose Prediction for Diabetic People}

\author{Maxime~De~Bois,
        Moun\^{i}m~A.~El~Yacoubi,
        and~Mehdi~Ammi
\thanks{M. De Bois is with CNRS-LIMSI and the Universit\'{e} Paris-Saclay, Orsay, France (e-mail: maxime.debois@limsi.fr).}
\thanks{M. A. El Yacoubi is with Samovar, CNRS, T{\'e}l{\'e}com SudParis, Institut Polytechnique de Paris, \'{E}vry, France}
\thanks{M. Ammi is with Universit\'{e} Paris 8, Saint-Denis, France}}


\maketitle

\IEEEpeerreviewmaketitle

\begin{abstract}
Deep learning has yet to revolutionize general practices in healthcare, despite promising results for some specific tasks. This is partly due to data being in insufficient quantities hurting the training of the models. To address this issue, data from multiple health actors or patients could be combined by capitalizing on their heterogeneity through the use of transfer learning. 

To improve the quality of the transfer between multiple sources of data, we propose a multi-source adversarial transfer learning framework that enables the learning of a feature representation that is similar across the sources, and thus more general and more easily transferable. We apply this idea to glucose forecasting for diabetic people using a fully convolutional neural network. The evaluation is done by exploring various transfer scenarios with three datasets characterized by their high inter and intra variability. 

While transferring knowledge is beneficial in general, we show that the statistical and clinical accuracies can be further improved by using of the adversarial training methodology, surpassing the current state-of-the-art results. In particular, it shines when using data from different datasets, or when there is too little data in an intra-dataset situation. To understand the behavior of the models, we analyze the learnt feature representations and propose a new metric in this regard. Contrary to a standard transfer, the adversarial transfer does not discriminate the patients and datasets, helping the learning of a more general feature representation. 

The adversarial training framework improves the learning of a general feature representation in a multi-source environment, enhancing the knowledge transfer to an unseen target. 

The proposed method can help improve the efficiency of data shared by different health actors in the training of deep models.

\end{abstract}

\section{Introduction}

Driven by the momentum created by recent successes in image recognition \cite{simonyan2014very} or natural language processing \cite{devlin2018bert}, the application of deep learning to the medical field is showing promising results (e.g., detection of diabetes retinopathy \cite{gulshan2016development}, skin cancer classification \cite{esteva2017dermatologist}, or breast cancer detection \cite{wang2016deep}). However, deep learning has yet to revolutionize healthcare practices, for which its application is facing several challenges \cite{ching2018opportunities}. Whereas some of them are linked to the nature of the deep models themselves, with, for instance, the interpretability or interoperability of the models, other challenges are related to the data. Indeed, data are needed in huge quantities for the models based on deep learning to succeed at their task. While most of the successful applications owe their success to data that have been acquired throughout the years (e.g., cancer images), in general, it is hard to obtain health-related data in sufficient quantities. This is due to the cost of their labeling which requires expert knowledge, to their sensitive nature making their sharing between healthcare structures difficult, and to the heterogeneity of data (e.g., different hardware, phenotypes, standards) complicating their simultaneous use \cite{wang2019deep}. 

To alleviate the lack of available data, different strategies can be considered. First, the original data can be artificially augmented by operating basic data transformations or by using data simulation \cite{dhungel2017deep}. Alternatively, the efficiency of the data can be increased, with for instance few-shot learning methodologies \cite{altae2017low}. Finally, prior knowledge can be instilled into the deep models in order to reduce the quantity of data needed for their training. This knowledge can either be domain-specific, expert knowledge \cite{holzinger2016interactive}, or it can be obtained by first training the model on other semi-related data and then finetuning it to the data of interest, which is known as transfer learning \cite{bar2015deep}. 

Transfer learning is especially interesting in the medical field because of the variety of the available sources. For instance, we can transfer knowledge between multiple hospitals or electronic health records, between datasets with different experimental settings, or even between patients in the case of personalized medicine. Furthermore, this situation opens the way to the combination of multiple sources for the extraction of knowledge, known as multi-source transfer learning \cite{christodoulidis2016multisource}. It enables the use of more data, which might not be in sufficient quantities in each individual source. Furthermore, it offers the opportunity to make the extracted knowledge more general and thus more easily transferable. However, the efficiency of the transfer heavily depends on the similarity between the source tasks and the target task, as a high dissimilarity has been shown to be a major contributor of negative transfer \cite{yosinski2014transferable}. There is also the risk of having the model learn how to discriminate data from different sources, and therefore hurting the generalization of the learnt model.

Ganin \textit{et al.} addressed this problem in the context of domain adaptation by proposing a domain-adversarial training methodology \cite{ganin2016domain}. Domain adaptation is a subfield of transfer learning and differs from general transfer learning (and therefore from multi-source transfer learning) by having the model trained on the source (labeled) and target (unlabeled) data conjointly. In this setting, the domain-adversarial training implements a module that discriminates the source domain from the target domain of the extracted features, and a feature extractor module working against that objective. This has the consequence of promoting the learning of a unique feature representation shared by both the source and the target. Inspired by their work, we propose to adapt the idea to multi-source transfer learning in order to improve the generalizability of the feature representation that is learnt on the source data. 


Our contributions can be summed up as follow: \begin{itemize}
    \item We first propose the multi-source adversarial transfer learning framework, to which we refer as \textit{adversarial transfer learning} (ATL) (to ease the reading, the \textit{multi-source} label is dropped when deemed not necessary). Deep models used traditionally in transfer learning (e.g., convolutional neural networks) generally implement two modules: a feature extractor that extracts knowledge from the inputs, and a predictor that uses the knowledge to make the predictions. In the ATL setting, a new module infers the source of the input data based on its extracted features. By making the features extractor compete against this objective, the learnt feature representation generalizes better across the sources. Our hypothesis is that the feature representation, being more general, will then transfer better to an unknown target. This idea is particularly well suited for healthcare because of the heterogeneity of the data making the simultaneous use of data with different origins difficult. Besides, compared to data that are usually hard to share, being sensitive and personal, a model can easily be shared between health actors, as it preserves the anonymity of the source data. Finally, after it has been shared, the model can easily be finetuned to the data of interest as it requires smaller data quantities. 
    \item We demonstrate the efficiency of the proposed method by applying it to the challenging task of glucose prediction for diabetic people. In glucose prediction, because the models are personalized to the patient, and because the data are very costly, we often do not have enough data for the training of deep models. However, data from other patients could be used to help the learning of a better personalized predictive model. To our knowledge, we are the first to propose the use of transfer learning of deep models in the field of glucose prediction.
    \item We investigate the transferability of the models when varying the source data. In this study, we use three significantly different datasets, the first one consisting of 6 type-2 diabetic patients, the second one comprising 6 type-1 diabetic patients, and the third being made of 10 \textit{in-silico} type-1 diabetic patients. We explore the transferability of the models in intra-dataset and inter-dataset settings as well as when using synthetic data or combinations of the aforementioned datasets to promote the learning of a general feature representation.
    \item We analyze the learnt feature representations and we propose, to this end, a new metric we call Local Domain Perplexity (LDP) that aims at quantifying the generalizability of the extracted features in a multi-source setting by looking at the distance between the extracted features of different sources.
    \item Finally, we released the source code of the study as well as the weights of the models in our GitHub repository \cite{debois2020atl}.
\end{itemize}{}
      
Our paper is structured as follows. First, we give a more formal definition of what multi-source transfer learning is and introduce the proposed method. Then, we review the many facets and challenges of the glucose prediction problem and how the proposed methodology can help to address them. We then provide the reader with extensive experimental details, from the datasets we used and their preprocessing, to the building of the models and their evaluation. Finally, we compare the results of the standard and adverse multi-source transfer methodologies in every possible transfer scenario and analyze the results. By significantly improving upon state-of-the-art results in the field, the multi-source adversarial transfer learning framework is shown to address efficiently the data shortage issue in the biomedical field.

\section{Multi-Source Adversarial Transfer Learning}

\begin{figure*}
    \centering
    \includegraphics[width=\linewidth]{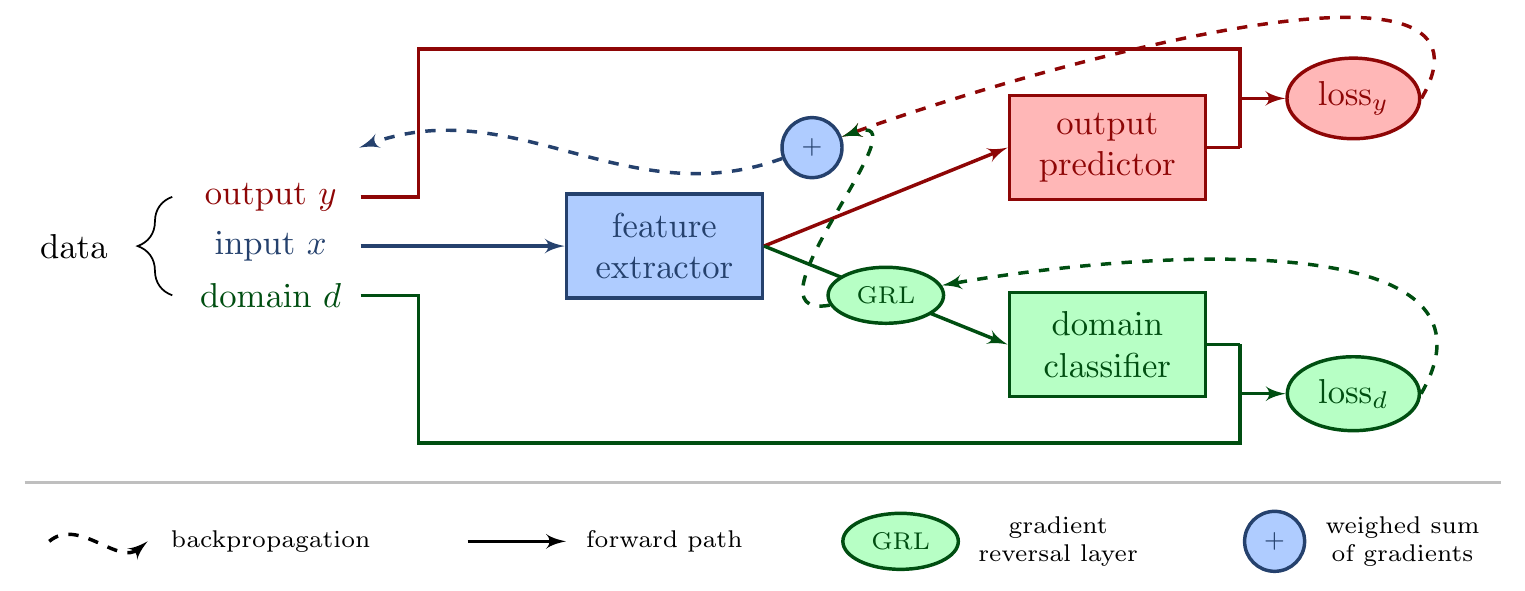}
    \caption{General representation of a deep model trained with the adversarial training methodology in the multi-source transfer learning setting. The model is made of three different parts: a \textit{feature extractor}, an \textit{output predictor}, and a \textit{domain classifier}. The adversarial nature of the learning is achieved by having the feature extractor compete against the domain classifier thanks to the reversal of $\text{loss}_d$ (multiplication by $-1$) when passing through the gradient reversal layer (GRL) during backpropagation.}
    \label{fig:dann_general}
\end{figure*}

The goal of this section is to give a formal definition of multi-source transfer learning, to highlight the challenges of its application in healthcare, and then to describe the proposed multi-source adversarial transfer learning methodology that aims at addressing them.

\subsection{Transfer Learning}

A \textit{domain} $\mathcal{D}$ is defined by a feature space $\mathcal{X}$ and a marginal probability distribution $P(X)$, where $X \in \mathcal{X}$. A \textit{task} $\mathcal{T}$ consists in an objective space $\mathcal{Y}$ and an objective predictive function $f(\cdot)$. $f(\cdot)$ is unknown but can be learnt from data samples $\{x_i,y_i\}$, where $x_i \in X$ and $y_i \in \mathcal{Y}$.

Pan and Yang defined \textit{transfer learning} as follows \cite{pan2009survey}: given a source domain $\mathcal{D}_S$ and learning task $\mathcal{T}_S$, a target domain $\mathcal{D}_T$ and learning task $\mathcal{T}_T$, \textit{transfer learning} $\{\mathcal{D}_S,\mathcal{T}_S\}\rightarrow\{\mathcal{D}_T,\mathcal{T}_T\}$ aims at improving the learning of the target predictive function $f_T(\cdot)$ in $\mathcal{D}_T$ using the knowledge in $\mathcal{D}_S$ on $\mathcal{T}_S$, where $\mathcal{D}_S \neq \mathcal{D}_T$ or $\mathcal{T}_S \neq \mathcal{T}_T$.

In this paper, we will focus on \textit{inductive} transfer learning, which is the most common type of transfer learning. Inductive transfer learning is characterized by having closely related source and target tasks, and by having some labeled data in the target domain and enough labeled data in the source domain. Using deep models, this kind of transfer is usually done by training a first model on the source domain, and then by finetuning it (or a portion of it) on the target domain. 

\subsection{Multi-Source Transfer Learning}

In \textit{multi-source} transfer learning, the knowledge we aim at transferring is conjointly learnt on several source pairs $\{\mathcal{D}_{S_n},\mathcal{T}_{S_n}\}$, each of them being different from one another and different from the target $\{\mathcal{D}_T, \mathcal{T}_T\}$. By learning on several source domains, we aim at addressing a potential data quantity issue in the invidual sources as well as at learning a feature representation that is more general, and thus hopefully more useful when being transferred to the target domain.

The nature of the source domains can vary a lot, requiring us to be careful when selecting them, in order to avoid scenarios of negative transfers (a transfer that harms the learning of the target task in the target domain, instead of helping it). This is especially true in the medical field where data are heterogenous, having different probability distributions due to their origin (e.g., different patients, sensors, collection environments), different formats (e.g., image resolution, sample frequency), different scales (e.g., color scale for images, units for physical or physiological measurements), or simply being in different quantities \cite{wang2019deep}. 

To make a positive transfer possible, these differences need to be addressed either before or  during the training of the model we want to transfer. Differences in scale or format can easily be eliminated through the transformation of the data beforehand (rescaling, reshaping, standardization). Uneven data quantities across the domains can be solved the same way as imbalanced datasets in classification problems with, for instance, sample-reweighing or data-augmentation techniques \cite{kotsiantis2006handling}. On the other hand, the difference in probability distributions could be healthy for the building of a general model, taking advantage of the diversity of the data. However, there is a risk, for the model, of learning how to discriminate the different source domains and learn a distinct feature representation for each of them.  This kind of feature representation would be less general by being heavily specific, overfitted, to these individual domains, harming its future transfer to the unseen target domain.

\subsection{Multi-Source Adversarial Transfer Learning}

\label{sec:dann}

 To address the issue of the over-specialization of the models trained on the multiple sources, inspired by the work of Ganin \textit{et al.} in the field of domain adaptation, we propose the multi-source adversarial transfer learning framework. 


Fig. \ref{fig:dann_general} provides a general graphical representation of a model using the adversarial training methodology in a  multi-source transfer learning setting. The features computed by the feature extractor module are used by the output predictor and the domain classifier to respectively predict the output, and the domain the data come from. The output predictor and the domain classifier are both classically trained by backpropagating their respective losses, the first one depending on the problem the model aims at solving (is it a regression or classification problem?), the second one being the multi-class cross-entropy (each class representing one source domain). When arriving at the feature extractor module, the loss of the domain classifier is reversed (multiplied by $-1$) by the gradient reversal layer. As a consequence, while the feature extractor learns a feature representation that is useful for the output prediction, it also learns a feature representation that is indiscriminative of the domain the data come from, and thus promotes a more general one. The bias-vs-variance tradeoff of the generalization is balanced by tweaking a coefficient, $\lambda$, weighing the magnitude of the domain-classifier-related gradient. To cope with imbalanced datasets, the multi-class cross-entropy penalty associated to a given domain is further weighed inversely proportionally to its representation among the domains.

When finetuning the model on the target domain, the domain classifier has no purpose anymore and can be removed (either by transferring the feature extractor and output predictor weights into a new model that does not have a domain classifier, or by setting $\lambda$ to 0).

\section{Glucose Prediction for Diabetic People}


In this section, we review the state of the art of the glucose prediction field, aiming at highlighting the high diversity of the datasets used in the studies, the numerous challenges the field is facing, and how transfer learning can help in this regard.

Diabetic people suffer from the regulation of their glucose level, troubled by either the non-production of insulin (type-1 diabetes, T1D), or by an increasing body resistance to its action (type-2 diabetes, T2D). Whereas they face short-term consequences (e.g., clumsiness, coma, death) when their glucose level falls too low (hypoglycemia), the consequences are more long-term (e.g., cardiovascular diseases, blindness) when it gets too high (hyperglycemia). In order to avoid being in either state, the forecasting of future glucose values could warn the patient of its arrival, enabling him/her to take the appropriate actions.

The topic of glucose prediction for diabetic patients is not new. In 2007, Sparacino \textit{et al.} showed that the task was doable using autoregressive (AR) models \cite{sparacino2007glucose}. They used data coming from 28 T1D patients, whose glucose values have been collected for 48 hours through the use of a continuous glucose monitoring (CGM) device, the Glucoday. Following this lead, a lot of different kind of AR models have been tried out throughout the years (e.g., AR with exogenous inputs, ARX \cite{daskalaki2013early}; AR with moving average, ARMA \cite{eren2012adaptive}). 

However, we are currently witnessing a shift towards the use of more advanced machine learning models. This shift is made possible by the increasing availability of data using better CGM devices, which are more accurate and have a higher sampling frequency. Besides, those data quantities enable the building of personalized glucose predictive models which are more efficient, due to the high inter/intra variability of the diabetic population \cite{oviedo2017review}. In 2012, Georga \textit{et al.} explored the use of Support Vector Regression (SVR) on a cohort of 27 T1D patients for the forecasting of glucose at a prediction horizon (PH) ranging from 15 to 120 minutes \cite{georga2013multivariate}. In their study, the patients have been monitored for an average of 13.42 $\pm$ 3.69 days, during which glucose values (Guardian Real-Time CGM device, Medtronic Minimed, Inc.), energy expenditure (SenseWear Armband, BodyMedia, Inc.), carbohydrate (CHO) intakes, and insulin infusions (both using a paper diary) have been collected. In 2016, Jankovic \textit{et al.} proposed to use Gaussian processes (GP) for the prediction of current glucose values from manually recorded events and time (paper diary) collected on 10 T1D patients \cite{tomczak2016gaussian}.

Thanks to the recent progress in the field of deep learning, neural-network-based models are heavily investigated for the task of glucose prediction. In 2012, Daskalaki \textit{et al.} showed that feed-forward neural networks (FFNN) outperform standard AR and ARX models for a PH of 30-to-45 minutes   \cite{daskalaki2012real}. They used the Type 1 Diabetes Metabolic Simulator (T1DMS) \cite{man2014uva} to simulate 8 days of data of the software's 10-adults \textit{in-silico} population. Recurrent neural networks, and especially those based on LSTM units, have seen some recent use because of their inherent nature making them well suited for the forecasting of time-series, and thus, of future glucose values \cite{mirshekarian2017using, de2019prediction}. Following their success in the image recognition field, convolutional neural networks (CNN) have drawn interest to the glucose prediction community \cite{zhu2018deep, li2019glunet} as well. In particular, Zhu \textit{et al.} explored their use using the OhioT1DM dataset \cite{zhu2018deep}. This dataset, which has been made publicly available \cite{marling2018ohiot1dm}, is made of 6 T1D patients that have been monitored for 8 weeks and for which data related to glucose values, CHO intakes, insulin infusions, physical activity, skin conductance and temperature, have been collected.

Given the high diversity of the experimental settings, the high inter-/intra-subject variability, and given the complexity of collecting sufficient quantities of such sensitive data, transfer learning is a promising approach to alleviate this issue. That being said, it has not been explored much in the past in this field. In 2010, Gali \textit{et al.} showed that it is possible to build global glucose predictive models that work well when tested on unseen patients (no finetuning to the target patients) \cite{gani2009universal}. For that purpose, they used an ARX model. They did their analysis using three different datasets, all of them having different experimental settings. Whereas two of them were made of T1D patients, the other one was composed of T2D patients. This study suggests that those datasets share some common characteristics that can be transferred to new patients or new datasets. Recently, Luo \textit{et al.} proposed a methodology to transfer ARX models learnt on several patients to another, unseen, patient \cite{luo2019transfer}. They validated their results using the T1DMS software, with a simulation length of 6 days. They showed that their methodology outperforms individual models and enables the use of less data for the new patient. However, while promising, those results have been obtained on a synthetic dataset, requiring further investigations on real-world data; besides, ARX models have been shown to be outperformed by non-linear machine-learning-based models (e.g., SVR, GP) and deep models (e.g., FFNN, LSTM) \cite{debois2020glyfe}.


Following the advances in the field, the use of transfer learning for the forecasting of future glucose values for diabetic people raises questions we intend at answering in this study:
\begin{enumerate}
    \item Can we transfer knowledge between real-world diabetic patients, given the high inter-/intra-subject variability of the disease?
    \item Is transfer learning useful for deep-learning-based models?
    \item Can we transfer between patients whose data have been collected in different experimental settings (e.g., different sensors, environments)?
    \item Can we transfer between type-1 and type-2 diabetic patients?
    \item Can synthetic data be used for the transfer to real-world data?
\end{enumerate}
\section{Methods}

This section presents the whole methodology that has been followed throughout the study, from the acquisition and preprocessing of the experimental data, to the training and evaluation of the models.

\subsection{Experimental Data}

One of the goals of this study is to compare the transferability of the models when varying the source and target datasets. We present here three different datasets: the IDIAB dataset (I), the Ohio dataset (O), and the T1DMS dataset (T). We focus our attention on the major differences that exist between them (e.g., diabetes type, patient diversity, data shape and scale inconsistency). 

~
\subsubsection{IDIAB Dataset (I)} Collected by ourselves between 2018 and 2019 (ID RCB 2018-A00312-53), the IDIAB dataset is made of data coming from 6 T2D patients (5F/1M, age 56.5 $\pm$ 9.14 years old, BMI 33.52 $\pm$ 4.17 $kg/m^2$). The patients have been monitored for 31.17 $\pm$ 1.86 days in free-living conditions wearing FreeStyle Libre CGM devices (Abbott Diabetes Care) and using the mySugr coaching application for diabetes as a diary. Whereas glucose values (in $mg/dL$) have been obtained every 15 minutes on average, insulin infusion (in units) and CHO intakes (in $g$) have been collected every minute. When looking at the glucose signals, numerous erroneous values have been identified (characterized by high-amplitude spikes). We chose to remove those values as their presence would have hurt the training of the models \cite{de2019prediction}.

~
\subsubsection{OhioT1DM Dataset (O)} Made public for the Blood Glucose Level Prediction Challenge in 2018, the OhioT1DM dataset has been collected on 6 T1D patients \cite{marling2018ohiot1dm} (2M/4F, age between 40 and 60 years old, BMI not disclosed). The patients wore MiniMed\textsuperscript{\textregistered} 530G insulin pumps (Medtronic), Enlite\textsuperscript{\textregistered} CGM sensors (Medtronic), and Basis Peak fitness bands for a duration of 8 weeks in real-life environments. In this study, we solely use the glucose signals (in $mg/dL$), the insulin infusions (in units), and the CHO intakes (in $g$), sampled every 5 minutes, to remain consistent with the other datasets.

~
\subsubsection{T1DMS Dataset (T)} The T1DMS dataset is made of 10 \textit{in-silico} T1D adults. The data have been generated by the Type 1 Diabetes Metabolic Simulator (T1DMS) \cite{man2014uva}, approved by the FDA as a substitute for pre-clinical animal testing, and thus widely used in the glucose prediction community \cite{debois2020glyfe}. The virtual patients were subject to a 8-week-long open-loop simulation following a 3-meal daily scenario. At the start of every meal, an insulin infusion is taken by the patient. More details of the simulation's scenario can be found in our previous work \cite{debois2020glyfe}. In the end, the glucose values (in $mg/dL$), the insulin infusion (in $pmol$), the CHO intakes (in $g/min$ for the duration of the meal which lasted 15 minutes) have been collected with a sample every minute.

~
\subsection{Preprocessing}

The objective of the preprocessing stage is to prepare the data for the training, and evaluation of the deep models. 
The preprocessing stage has two different objectives. The first one is to clean the data, getting rid of erroneous values that would prevent the models from learning well. Due to the heterogeneity of the datasets, the amount and type of cleaning vary from one to another. Then, in order to make the models interoperable between datasets, the data of every dataset need to have the same shape and scale. We transformed the IDIAB and T1DMS dataset to have the same units and sample frequency as the Ohio dataset. Finally, we prepare the datasets for the cross-validated evaluation of the models. Overall, more details about the preprocessing steps can be found in our previous work \cite{debois2020glyfe}. 


~
\subsubsection{Reshaping and Rescaling} In order to have models interoperable between datasets, the data need to have the same shape and scale across them. 
Here, while the IDIAB and Ohio datasets use the same units, the T1DMS dataset, in particular for the insulin infusions and CHO intakes, does not. The T1DMS' insulin infusions have been divided by 6000 to convert the values from $pmol$ to insulin units. The CHO intakes of every meal have been accumulated and timed at the start of the meal. 
Furthermore, the three datasets have a different sampling frequency. We resampled the IDIAB and T1DMS datasets to one sample every 5 minutes (sampling frequency of the Ohio dataset), resulting in an upsampling and a downsampling respectively. The downsampling of the T1DMS dataset has been done by taking the mean of the glucose signals and by summing up the insulin infusions and CHO intakes. On the other hand, the upsampling of the IDIAB data introduced a lot of missing values. Some of those values can be recovered following this strategy: linearly interpolate missing values if they are surrounded by known values, linearly extrapolate if not. 


~
\subsubsection{Samples Creation} The data samples used in the training and evaluation of the models are created by using the 3-hour history of glucose, insulin, and CHO values, as well as the 30 minutes ahead glucose value (the prediction ground truth). When the ground truth is not known, the sample is discarded to prevent the model from training on artificial data.

~
\subsubsection{Splitting}The OhioT1DM dataset is originally split into training and testing sets with the last 10 days of a given patient forming the testing set, and the remaining the training set. Similarly, we split the T1DMS and the IDIAB datasets having the last 10 and 5 days as the testing sets respectively. While the T1DMS dataset has roughly the same amount of data as the OhioT1DM dataset, this is not the case for the IDIAB dataset which has around half its amount.

Every training set is then split into a training and a validation set following a 80\%/20\% distribution. Whereas the goal of the training set is to train the models, the validation set is used as a prior evaluation of the model when optimizing its hyperparameters. This ensures that no data from the testing sets is used when building the models.

~
\subsubsection{Standardization}

Every patient's data signals have been standardized (zero mean and unit variance) w.r.t. their training set. Standardizing the data is a mandatory step for the training of neural-network-based models. It  has the advantage of also making the data of different patients more similar, which should help the building of more general, and thus better transferable models.

\subsection{Models}

\begin{figure*}
    \centering
    \includegraphics[width=\linewidth]{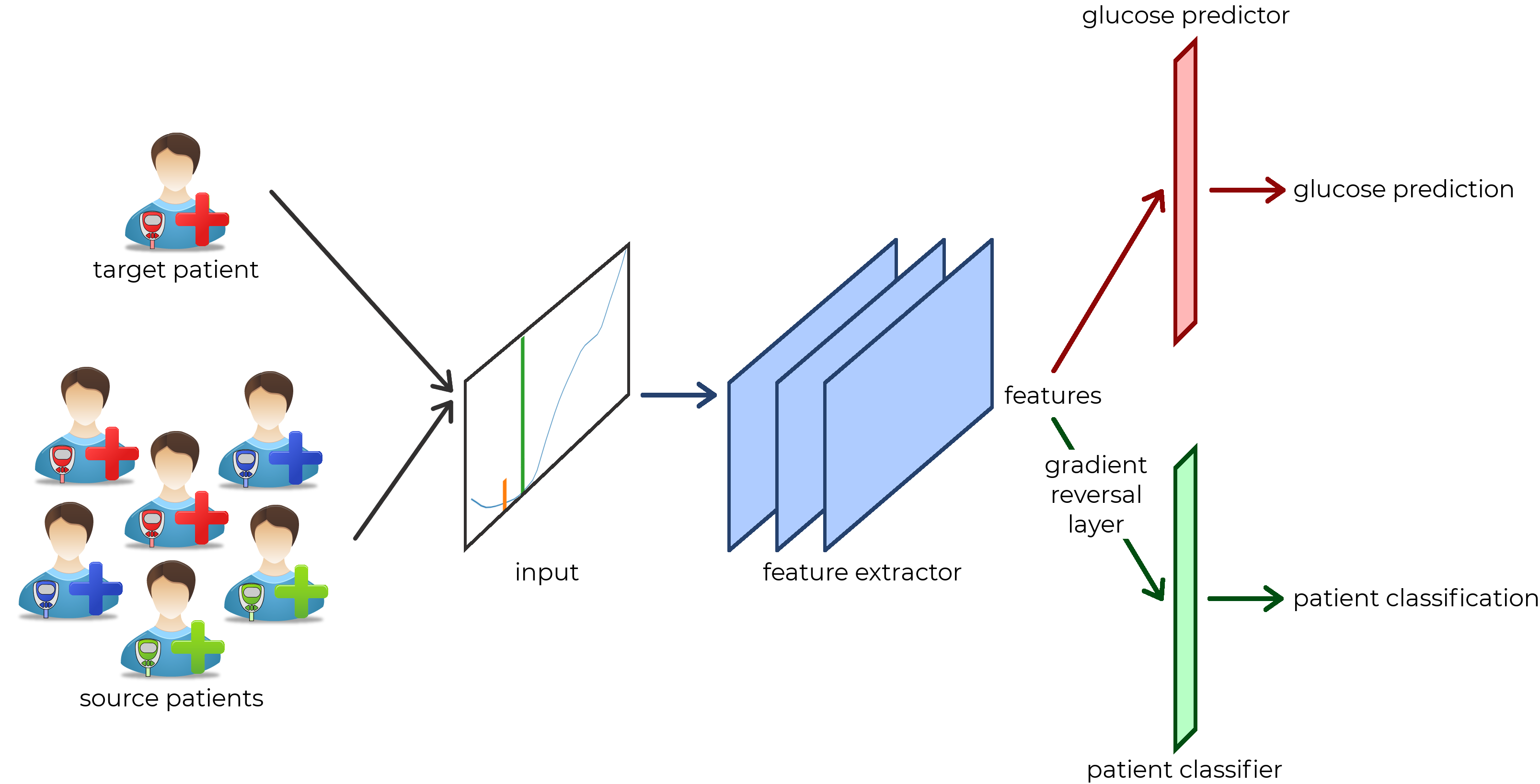}
    \caption{CNN-based adversarial transfer learning model for glucose prediction. A first model is trained on source patients that may come from different datasets and is then finetuned to the target patient. Thanks to the gradient reversal layer, the patient classifier module makes the feature extractor learn a feature representation that is general across the source patients.}
    \label{fig:atl_glucose}
\end{figure*}{}

We present here the models that we have used in this study, most of which being based on fully convolutional neural networks (FCN). The source code is available in our GitHub repository \cite{debois2020atl}. Furthermore, we also released the weights of the models trained on the source patients (before their finetuning to the target patient), so they can be used by anyone to build better glucose predictive models.

~
\subsubsection{Baseline Models}

In this study, we use three baseline models: a SVR model, and two FCN models, namely FCN \#1 and FCN \#2. The three models are solely trained and evaluated on the target patients.


The SVR model is based on our benchmark study \cite{debois2020glyfe}, in which it stands out from the other models, and in particular deep models, by being the overall best model for glucose prediction. As the overall methodology is identicaly, it is considered as our gold standard. Using a model that is not based on deep learning for comparison is particularly interesting as it can show, whether or not, the proposed approach enables deep models to surpass traditional machine-learning-based ones.

As for the FCN \#1 and \#2 models, while they share the same architecture, they differ in their training hyperparameters. They are made of 2 modules: a features extractor and a glucose predictor. Whereas the features extractor module is made of 3 layers (1-dimensional convolution of size 3 $\rightarrow$ ReLU activation function $\rightarrow$ batch normalization $\rightarrow$ dropout), with 64, 128, and 64 channels respectively, the predictor module is made of a single convolution of 2048 channels and of size 30 (making it behave like a fully connected layer). 
While both baseline models use the Adam optimizer to minimize the mean-squared error (MSE) loss function with an early stopping patience of $250$ epochs, FCN \#1 uses a learning rate of $10^{-4}$ and a dropout rate of $50\%$, and FCN \#2 uses a learning rate of $10^{-3}$ and a dropout rate of $90\%$. 
The reason behind using both architectures as baselines is that, while FCN \#1 shares the same hyperparameters as the models used for transfer (see below), the performances were not as good as expected because of the weak regularization given the amount of available data without transfer. By using FCN \#2, we make sure that our FCN baseline is strong enough.

~
\subsubsection{Standard Transfer Models}

We call standard transfer learning models (TL), the models that have been trained on source patients (global TL) and then finetuned to the target patient (finetuned TL). All the TL models are identical to FCN \#1. During finetuning, the learning rate and the early stopping patience have been decreased to $10^{-5}$ and $50$ epochs respectively. 

~
\subsubsection{Adversarial Transfer Models}

The adversarial transfer learning models (ATL, both global and finetuned) and the TL models share the exact same architecture and training methodology, at the exception of the presence of the domain classifier module and the gradient reversal layer in the former (see Section \ref{sec:dann}). The domain classifier is made of a single convolution of 2048 channels and a size of 30 (identical to its predictor module). It minimizes the multi-class cross-entropy weighed by $\lambda=10^{-0.75}$ with the Adam optimizer. The value of $\lambda$ has been chosen among 9 values between $10^{-2}$ and $10^1$ in a logarithmic scale \cite{ganin2016domain}. Moreover, to account for the imbalanced representation of the patients in the training sets, the loss associated with samples from a given patient are  weighed inversely proportionally to the total number of samples of this patient.

Fig. \ref{fig:atl_glucose} provides a graphical representation of the ATL model.



\subsection{Evaluation}

The models have been evaluated with a 5-fold cross-validation on the target subject. The results are averaged over the cross-validation permutations and over the target dataset (the standard deviation is computed on the patients' average results).

Two metrics are used to measure the accuracy of the models: the root-mean-squared error (RMSE), and the mean percentage absolute error (MAPE). Whereas the RMSE provides a real-scale measure of the average error, the MAPE is less sensitive to data distribution between the patients in the target domain.

Besides, we evaluate the clinical acceptability of the models with the Point-Error Grid Analysis (P-EGA), also known as the Clarke Error Grid \cite{clarke1987evaluating}. Given the known ground truth, a glucose prediction is given a mark from A to E depending on its clinical accuracy. Whereas an A prediction means the prediction is clinically accurate, an E prediction is a life-threatening prediction. The number of predictions that have either an A or B mark (A+B) is also often reported, as those predictions a deemed to have an acceptable clinical accuracy.

In the analysis of the results, we provide the significance tests of the performance ratio of one model, $M_2$, over another reference model $M_1$. We report the 99\% confidence interval (CI) of the paired ratio between $M_2$ and $M_1$ in MAPE. With such CI, the performance ratios shall fall outside the intervals only one time out of 100. The analysis of the intervals themselves depends on whether the chosen metric needs to be minimized or maximized. For instance, when it needs to be minimized (e.g., RMSE, MAPE), if the interval comprises 1, the results are not significant. However, if it does not, the results are significant, an interval below 1 denoting the significant improvement of $M_2$ over $M_1$, and an interval above 1 denoting the significant deterioration of $M_2$ over $M_1$.

To ease the analysis of the results, we will consider de following transfer scenario groups:
\begin{itemize}
    \item \textit{intra}: the source and target patients belong to the same dataset (i.e., I$\rightarrow$I, O$\rightarrow$O);
    \item \textit{inter}: the source and target patients do not belong to the same dataset and the source patients are not virtual (i.e., O$\rightarrow$I, I$\rightarrow$O);
    \item \textit{synth}: the source patients come from a virtual dataset (i.e., T$\rightarrow$I, T$\rightarrow$O);
    \item and any combination of these single-dataset scenarios (e.g., \textit{intra+inter} denoting IO$\rightarrow$I and IO$\rightarrow$O).
\end{itemize}
\section{Experimental Results}

The experimental numerical results are described by Table \ref{table:res_baseline}, Table \ref{table:res_ft} and \ref{table:atl_pega} representing, respectively, the accuracy of the baseline models, the accuracy of the global and finetuned models using transfer learning, and the clinical accuracy of all of them.

\begin{table}[]
\centering
\footnotesize
\caption{
Accuracy of the baseline models with mean (standard deviation) for both IDIAB and OhioT1DM populations.}
\label{table:res_baseline}
\begin{tabularx}{\linewidth}{C||C|C|C}
\toprule

\textbf{Dataset} & \textbf{Model}  & \textbf{RMSE} &\textbf{MAPE}  \\

\midrule

\multirow{3}{*}{IDIAB} & SVR & \underline{\textbf{20.20 \scriptsize{(5.90)}}} & \underline{\textbf{8.76 \scriptsize{(0.44)}}} \\
& FCN \#1 & 21.06 \scriptsize{(5.14)} & 9.66 \scriptsize{(1.00)} \\
& FCN \#2 & 20.64 \scriptsize{(5.20)} & 9.62 \scriptsize{(1.27)} \\

\midrule

\multirow{3}{*}{OhioT1DM} & SVR * & \underline{\textbf{20.10 \scriptsize{(2.34)}}} & \underline{\textbf{9.08 \scriptsize{(2.12)}}} \\
& FCN \#1 & 21.51 \scriptsize{(1.89)} & 9.82 \scriptsize{(2.08)} \\
& FCN \#2 & 20.61 \scriptsize{(2.09)} & 9.34 \scriptsize{(2.07)} \\

\bottomrule
\end{tabularx}

        \begin{flushright}
        \footnotesize{* These results have been presented in our previous work \cite{debois2020glyfe}}
        \end{flushright}
\end{table}{}

\newcolumntype{Z}[1]{>{\centering}m{#1}}

\begin{table*}
    \footnotesize
    \caption{Accuracy of the global and finetuned models after transfer for every Source (S) / Target (T) combinations with mean (standard deviation), averaged on the target population.}
    \label{table:res_ft}
    \begin{tabularx}{\linewidth}{c|c||C|C|C|C||C|C|C|C}
            \toprule
            
            \multicolumn{2}{c||}{\multirow{2}{*}{\textbf{Dataset}}} & \multicolumn{2}{c|}{\textbf{Standard Transfer}} & \multicolumn{2}{c||}{\textbf{Adversarial Transfer}} & \multicolumn{2}{c|}{\textbf{Standard Transfer}} & \multicolumn{2}{c}{\textbf{Adversarial Transfer}}\\
            \multicolumn{2}{c||}{} & \multicolumn{2}{c|}{\textbf{(global)}} & \multicolumn{2}{c||}{\textbf{(global)}} & \multicolumn{2}{c|}{\textbf{(finetuned)}} & \multicolumn{2}{c}{\textbf{(finetuned)}} \\
            
            \midrule
            
            \textbf{S} & \textbf{T} & \textbf{RMSE} & \textbf{MAPE} & \textbf{RMSE} & \textbf{MAPE} & \textbf{RMSE} & \textbf{MAPE} & \textbf{RMSE} & \textbf{MAPE}\\
            
            \midrule
            
            I & \multirow{7}{*}{I} & 21.47 \scriptsize{(7.50)} & 9.67 \scriptsize{(1.48)} & \underline{\textbf{19.61 \scriptsize{(6.27)}}} & 8.95 \scriptsize{(1.00)} & 20.25 \scriptsize{(5.02)} & 8.96 \scriptsize{(1.50)} & \underline{\textbf{18.51 \scriptsize{(5.48)}}} & \underline{\textbf{8.44 \scriptsize{(1.07)}}} \\
            O &  & 21.70 \scriptsize{(5.75)} & 10.22 \scriptsize{(1.85)} & 19.87 \scriptsize{(6.01)} & 9.01 \scriptsize{(1.52)} & 19.26 \scriptsize{(4.97)} & 9.13 \scriptsize{(1.26)} & 18.84 \scriptsize{(5.75)} & 8.57 \scriptsize{(1.11)}\\
            T &  & 25.47 \scriptsize{(6.00)} & 11.11 \scriptsize{(1.60)} & 29.57 \scriptsize{(6.01)} & 14.53 \scriptsize{(3.14)} & 20.08 \scriptsize{(4.94)} & 9.26 \scriptsize{(0.85)} & 19.50 \scriptsize{(5.14)} & 9.02 \scriptsize{(1.16)}\\
            IO &  & \underline{\textit{20.20 \scriptsize{(5.90)}}} & 9.51 \scriptsize{(1.49)} & 19.66 \scriptsize{(6.48)} & \underline{\textbf{8.90 \scriptsize{(1.52)}}} & \underline{\textit{19.10 \scriptsize{(5.04)}}} & \underline{\textit{8.95 \scriptsize{(1.00)}}} & 18.75 \scriptsize{(6.01)} & 8.50 \scriptsize{(1.23)}\\
            IT &  & 22.25 \scriptsize{(8.28)} & 10.61 \scriptsize{(2.91)} & 22.96 \scriptsize{(8.22)} & 10.52 \scriptsize{(1.98)} & 19.45 \scriptsize{(5.08)} & 9.03 \scriptsize{(1.16)} & 19.70 \scriptsize{(6.21)} & 8.93 \scriptsize{(1.10)}\\
            OT &  & 22.25 \scriptsize{(8.28)} & 10.61 \scriptsize{(2.91)} & 23.16 \scriptsize{(6.44)} & 10.20 \scriptsize{(1.61)} & 19.45 \scriptsize{(5.31)} & 9.04 \scriptsize{(1.20)} & 19.47 \scriptsize{(6.60)} & 8.79 \scriptsize{(1.13)}\\
            IOT &  & 20.72 \scriptsize{(6.34)} & \underline{\textit{9.45 \scriptsize{(2.02)}}} & 22.46 \scriptsize{(9.40)} & 9.87 \scriptsize{(2.07)} & 19.44 \scriptsize{(5.24)} & 9.02 \scriptsize{(1.07)} & 18.79 \scriptsize{(5.82)} & 8.59 \scriptsize{(1.05)}\\
            
            \midrule
            
            I  & \multirow{7}{*}{O} &  24.01 \scriptsize{(2.24)} & 11.62 \scriptsize{(1.85)} & 21.45 \scriptsize{(1.50)} & 10.15 \scriptsize{(1.70)}& 20.52 \scriptsize{(2.08)} & 9.49 \scriptsize{(2.18)} & 19.74 \scriptsize{(2.13)} & 8.96 \scriptsize{(2.02)}\\
            O & & 21.95 \scriptsize{(1.98)} & 10.20 \scriptsize{(2.10)} & 20.22 \scriptsize{(1.48)} & 9.18 \scriptsize{(1.83)} & 19.92 \scriptsize{(2.02)} & 9.09 \scriptsize{(2.14)} & 19.27 \scriptsize{(1.78)} & 8.68 \scriptsize{(1.97)}\\
            T &  & 30.17 \scriptsize{(4.64)} & 14.18 \scriptsize{(4.30)} & 36.63 \scriptsize{(7.99)} & 18.16 \scriptsize{(5.35)} & 20.20 \scriptsize{(1.99)} & 9.20 \scriptsize{(2.03)} & 19.93 \scriptsize{(1.74)} & 9.13 \scriptsize{(1.87)}\\
            IO &  &  21.17 \scriptsize{(2.16)} & 9.81 \scriptsize{(2.04)}& 19.58 \scriptsize{(1.65)} & \underline{\textbf{9.04 \scriptsize{(2.10)}}} & 19.91 \scriptsize{(2.01)} & 9.06 \scriptsize{(2.08)} & \underline{\textbf{18.94 \scriptsize{(1.66)}}} & \underline{\textbf{8.50 \scriptsize{(1.87)}}}\\
            IT & &   23.46 \scriptsize{(2.60)} & 11.58 \scriptsize{(2.75)} & 26.44 \scriptsize{(5.25)} & 13.37 \scriptsize{(2.86)} & 20.03 \scriptsize{(1.88)} & 9.25 \scriptsize{(2.08)} & 19.57 \scriptsize{(2.02)} & 8.81 \scriptsize{(1.85)}\\
            OT & &   21.39 \scriptsize{(2.16)} & 9.72 \scriptsize{(2.16)} & 19.88 \scriptsize{(1.26)} & 9.36 \scriptsize{(1.55)} & 19.72 \scriptsize{(2.04)} & 8.97 \scriptsize{(2.18)} & 19.16 \scriptsize{(1.73)} & 8.64 \scriptsize{(1.94)}\\
            IOT & &  \underline{\textit{20.68 \scriptsize{(2.12)}}} & \underline{\textit{9.58 \scriptsize{(2.14)}}} & \underline{\textbf{19.45 \scriptsize{(1.78)}}} & 9.19 \scriptsize{(1.91)} & \underline{\textit{19.57 \scriptsize{(2.02)}}} & \underline{\textit{8.93 \scriptsize{(2.13)}}} & 18.99 \scriptsize{(1.72)} & 8.56 \scriptsize{(1.89)} \\
            
            \bottomrule
        \end{tabularx}
\end{table*}

\newcolumntype{Z}[1]{>{\centering}m{#1}}

\begin{table*}
    \footnotesize
    \caption{
    Clinical accuracy (P-EGA) of the baseline models and the best finetuned models after transfer for every Source (S) / Target (T) combinations with mean (standard deviation), averaged on the target population.}
    \label{table:atl_pega}
    \begin{tabularx}{\linewidth}{c|c||C|C|C|C|C|C}
            \toprule
            
            \multirow{2}{*}{\textbf{Dataset}} & \multirow{2}{*}{\textbf{Model}} & \multicolumn{6}{c}{\textbf{\textit{Point Error-Grid Analysis} (P-EGA)}}\\
            \cmidrule{3-8}
            & & \textbf{A+B} & \textbf{A} & \textbf{B} & \textbf{C} & \textbf{D} & \textbf{E} \\

            \midrule

            \multirow{6}{*}{IDIAB (I)} & SVR  & \underline{\textbf{99.18 \scriptsize{(0.43)}}} & 94.80 \scriptsize{(1.49)} & 4.38 \scriptsize{(1.33)} & 0.04 \scriptsize{(0.10)} & \underline{\textbf{0.78 \scriptsize{(0.49)}}} & \underline{\textbf{0.00 \scriptsize{(0.00)}}}\\
            & FCN \#1 & 98.43 \scriptsize{(1.58)} & 92.12 \scriptsize{(2.58)} & 6.32 \scriptsize{(1.62)} & \underline{\textbf{0.00 \scriptsize{(0.00)}}} & 1.57 \scriptsize{(1.58)} & \underline{\textbf{0.00 \scriptsize{(0.00)}}}\\
            & FCN \#2 & 98.21 \scriptsize{(1.71)} & 92.59 \scriptsize{(3.37)} & 5.61 \scriptsize{(2.09)} & \underline{\textbf{0.00 \scriptsize{(0.00)}}} & 1.79 \scriptsize{(1.71)} & \underline{\textbf{0.00 \scriptsize{(0.00)}}}
\\
            & TL * & 98.57 \scriptsize{(1.18)} & 93.73 \scriptsize{(2.68)} & 4.84 \scriptsize{(1.81)} & \underline{\textbf{0.00 \scriptsize{(0.00)}}} & 1.43 \scriptsize{(1.18)} & \underline{\textbf{0.00 \scriptsize{(0.00)}}}
\\
            & ATL $^\dagger$& 98.77 \scriptsize{(1.30)} & \underline{\textbf{94.96 \scriptsize{(2.59)}}} & \underline{\textbf{3.81 \scriptsize{(1.67)}}} & \underline{\textbf{0.00 \scriptsize{(0.00)}}} & 1.23 \scriptsize{(1.30)} & \underline{\textbf{0.00 \scriptsize{(0.00)}}}\\
            
            \midrule
            
            \multirow{6}{*}{OhioT1DM (O)} & SVR  & 99.10 \scriptsize{(0.88)} & 93.48 \scriptsize{(3.14)} & 5.62 \scriptsize{(2.33)} & 0.01 \scriptsize{(0.02)} & 0.88 \scriptsize{(0.87)} & 0.01 \scriptsize{(0.03)}\\
            & FCN \#1 & 98.67 \scriptsize{(1.18)} & 91.67 \scriptsize{(3.47)} & 7.01 \scriptsize{(2.38)} & \underline{\textbf{0.00 \scriptsize{(0.00)}}} & 1.33 \scriptsize{(1.18)} & \underline{\textbf{0.00 \scriptsize{(0.00)}}}\\
            & FCN \#2 & 98.67 \scriptsize{(1.07)} & 92.71 \scriptsize{(3.32)} & 5.96 \scriptsize{(2.29)} & \underline{\textbf{0.00 \scriptsize{(0.00)}}} & 1.33 \scriptsize{(1.07)} & \underline{\textbf{0.00 \scriptsize{(0.00)}}}\\
            & TL ** & 98.89 \scriptsize{(1.13)} & 93.69 \scriptsize{(3.63)} & 5.20 \scriptsize{(2.52)} & 0.00 \scriptsize{(0.01)} & 1.11 \scriptsize{(1.12)} & 0.00 \scriptsize{(0.01)}\\
            & ATL $^\ddagger$& \underline{\textbf{99.20 \scriptsize{(0.76)}}} & \underline{\textbf{94.44 \scriptsize{(2.83)}}} & \underline{\textbf{4.77 \scriptsize{(2.12)}}} & \underline{\textbf{0.00 \scriptsize{(0.00)}}} & \underline{\textbf{0.80 \scriptsize{(0.76)}}} & \underline{\textbf{0.00 \scriptsize{(0.00)}}}\\
            
            \bottomrule
        \end{tabularx}

        \begin{flushright}
        \footnotesize{* IO$\rightarrow$I, ** IOT$\rightarrow$I, $^\dagger$ I$\rightarrow$I, $^\ddagger$ IO$\rightarrow$O scenarios}
        \end{flushright}
\end{table*}

\begin{figure*}
        \centering
        \begin{subfigure}[b]{0.475\textwidth}
            \centering
            \includegraphics[width=\textwidth]{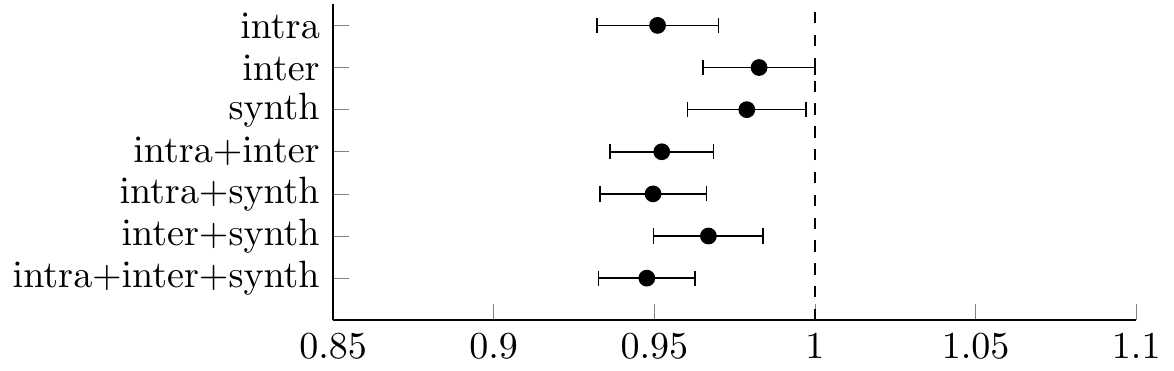}
            \caption[Network1]%
            {{\small Standard Transfer over FCN \#2}}    
            \label{fig:tl_vs_fcn2}
        \end{subfigure}
        \hfill
        \begin{subfigure}[b]{0.475\textwidth}   
            \centering 
            \includegraphics[width=\textwidth]{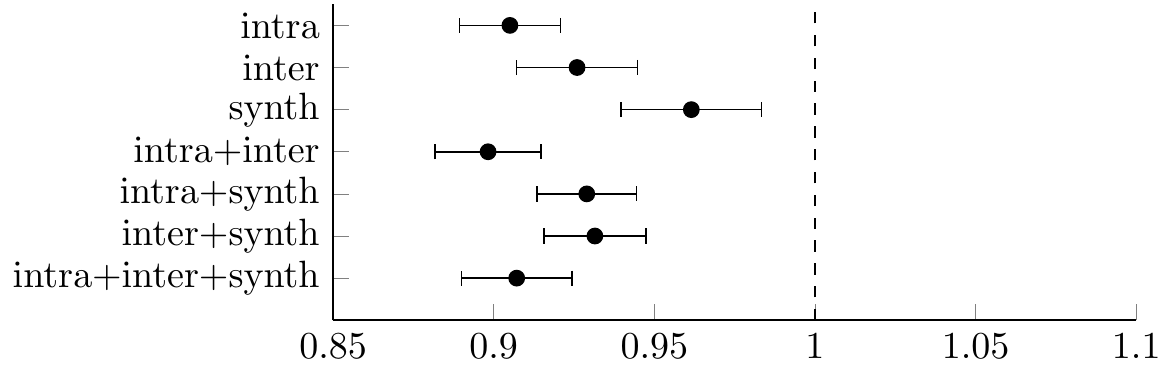}
            \caption[]%
            {{\small Adversarial Transfer over FCN \#2}}    
            \label{fig:atl_vs_fcn2}
        \end{subfigure}
        \vskip\baselineskip
        \begin{subfigure}[b]{0.475\textwidth}  
            \centering 
            \includegraphics[width=\textwidth]{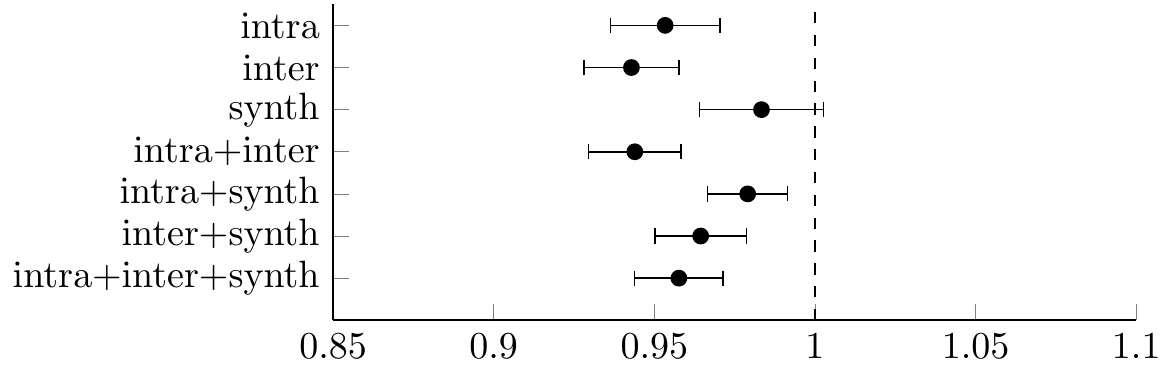}
            \caption[]%
            {{\small Adversarial Transfer over Standard Transfer}}    
            \label{fig:atl_vs_tl}
        \end{subfigure}
        \hfill
        \begin{subfigure}[b]{0.475\textwidth}   
            \centering 
            \includegraphics[width=\textwidth]{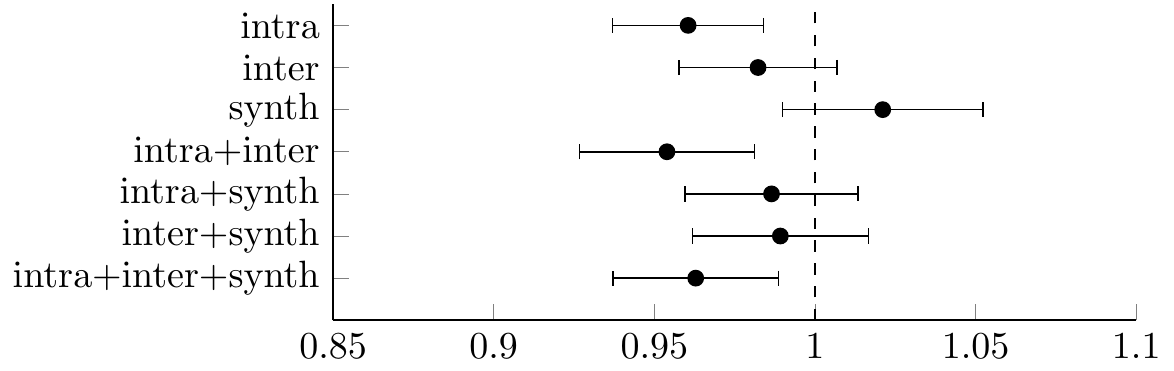}
            \caption[]%
            {{\small Adversarial Transfer over SVR}}    
            \label{fig:atl_vs_svr}
        \end{subfigure}
        \caption[ 99\% confidence intervals of the paired performance ratio in MAPE of one model over another for every possible transfer scenarios ]
        {\small 99\% confidence intervals of the paired performance ratio in MAPE of one model over another for every possible transfer scenarios.} 
        \label{fig:confidence_intervals}
\end{figure*}

\subsection{Results Analysis}

The baseline results displayed in Table \ref{table:res_baseline} are consistent with the findings in our benchmark study, where deep models are outclassed by standard machine-learning models, and in particular, by the SVR model \cite{debois2020glyfe}. While the added regularization in FCN \#2 improves upon the FCN \#1 results, it is still not enough. This suggests that data for a given patient are not in sufficient quantities.

The performances of the global TL models, displayed by Table \ref{table:res_ft}, are variable, depending on the source and target patients. First, all the scenarios that show a better accuracy than our reference model FCN \#2 use \textit {intra} data  (IO$\rightarrow$I, IOT$\rightarrow$I, O$\rightarrow$O, IO$\rightarrow$O, IOT$\rightarrow$O). Only the I$\rightarrow$I transfer does not have good results. This is explained by the lack of data within the IDIAB dataset, lack that is less important for the scenarios IO$\rightarrow$I or O$\rightarrow$O for example. Furthermore, using synthetic data seems to be effective only when they are used in combination with real data in sufficient quantities (i.e., when the source is IOT).

The use of the adversarial transfer learning methodology generally improves these initial global results. Additionally, it makes the I$\rightarrow$I transfer work. In addition, the scenario \textit{intra+inter} (IO$\rightarrow$I and IO$\rightarrow$O) stands out by having better and equivalent results than FCN \#2 and SVR respectively. However, we can note that the efficiency of the global transfer highly depends on the origin of the source and target patients. If no \textit{intra} patient is used as the source, then the global transfer is quite ineffective. Also, the use of additional synthetic data is less effective than for global TL models. Overall, the global ATL results suggest that the models make a better use of the data, even when present in low quantities, reducing the need of adding highly dissimilar data.

Compared to the standard global results, the  standard finetuned results show clear improvements for all transfer scenarios. In addition, these performances are significantly better than those obtained by our reference model FCN \#2 (see Figure \ref{fig:tl_vs_fcn2}). By finetuning the model to the target patient, we also enable the transfer from synthetic data to work. Overall, the best transfer scenarios are those that use \textit{intra}  (not exclusively). Overall, these results show the advantage of transferring knowledge between patients for the prediction of glucose. While using closely related data works better, the transfer can also work even when the data have different natures (type 1 or type 2, various experimental conditions or systems). We note, however, that some global ATL results are equivalent or better than those of finetuned TL models, suggesting a potential use of general models for predicting blood glucose using the multi-source adverse transfer training methodology without finetuning them to the target patient.

Using the adversarial methodology is showed to further improve upon the results obtained with the standard finetuned models, and this for almost all transfer scenarios (some scenarios, such as IT or OT$\rightarrow$I show equivalent results). As shown in Figure \ref{fig:atl_vs_tl}, the improvements related to the use of the adverse methodology are statistically significant and are particularly effective for the used real data (\textit{intra}, \textit{inter}, or both). Although the confidence interval of the \textit{synth} scenario prevent us from concluding on the significance of the improvement, we can appreciate the improvement by comparing the intervals of TL and ATL against FCN \#2 with Figures \ref{fig:tl_vs_fcn2} and \ref{fig:atl_vs_fcn2}. Finally, Figure \ref{fig:atl_vs_svr} indicates that the ATL results are significantly better than our gold standard SVR model for the transfer scenarios \textit{intra}, \textit{intra+inter} and \textit{intra+inter+synth}. This shows that transfer learning, and in particular the proposed adversarial multi-source methodology, alleviates the data issue deep models have and enables them to outperform state-of-the-art models based on standard machine learning.

Overall, even though transferring knowledge (standard or adverse) works for all the scenarios studied, its effectiveness is variable. Within the transfer scenarios using a single source dataset (\textit{intra}, \textit{inter} or \textit{synth}), the most efficient transfer is, not surprisingly, the transfer \textit{intra}. Adding new source data (e.g., \textit{intra+inter} compared to \textit{intra}, or \textit{inter+synth} compared to \textit{inter}) does not always improve the results. While it appears to be profitable in the case of a transfer to the OhioT1DM dataset, the best results for a transfer to IDIAB remain those of the scenario \textit{intra}. We hypothesize this is due to the lack of IDIAB data in comparison with the other two datasets. Augmenting the source IDIAB data with all the OhioT1DM or T1DMS data could drown the IDIAB data which remain the most important data for a transfer to another patient from the IDIAB dataset.

Table \ref{table:atl_pega} allows us to focus on the clinical acceptability of the results. In order to simplify its reading, we have chosen to only represent the clinical acceptability of the baseline models and the best (in MAPE) finetuned models in Table \ref{table:res_ft}. Unsurprisingly, the results obtained by ATL models show a clear improvement over our baselines FCN \#1 and \#2 as well as over TL models. Only the SVR model seems to remain competitive from the point of view of clinical acceptability. Indeed, it has particularly low prediction rates in the D and E regions, especially for the IDIAB dataset.

We can thus conclude on the importance of transfer learning, and in particular of the adverse transfer learning methodology for the training of glucose-predictive models based on deep learning. Although transferring knowledge is effective with data from different origins (real or simulated data, data from type 1 or 2 diabetics, or from various experimental conditions), it is preferable to use data \textit{intra}, data which can be augmented if necessary with \textit{inter} data. While the \textit{synth} transfer scenario is the least efficient here, it remains interesting in a situation of total or almost total absence of real data.

\subsection{Behavior Analysis}

\begin{figure*}
        \centering
        \begin{subfigure}[b]{0.475\textwidth}
            \centering
            \includegraphics[width=\textwidth]{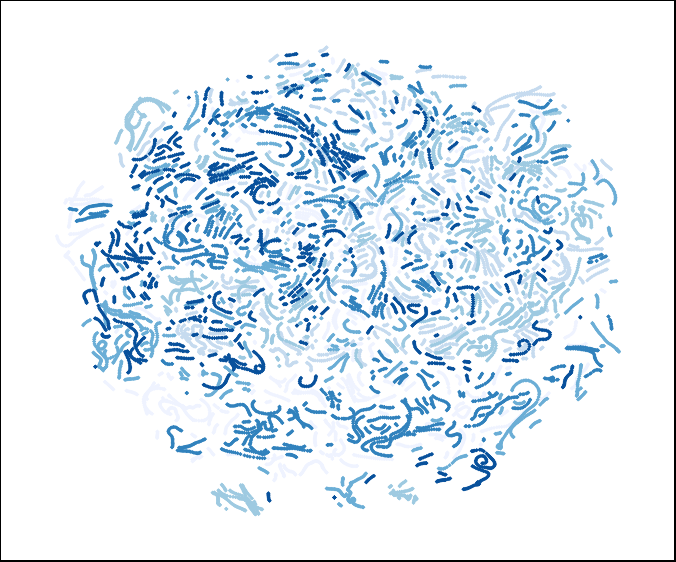}
            \caption[Network1]%
            {{\small O$\rightarrow$I : Standard Transfer}}    
            \label{fig:tsne_o2i_tl}
        \end{subfigure}
        \hfill
        \begin{subfigure}[b]{0.475\textwidth}  
            \centering 
            \includegraphics[width=\textwidth]{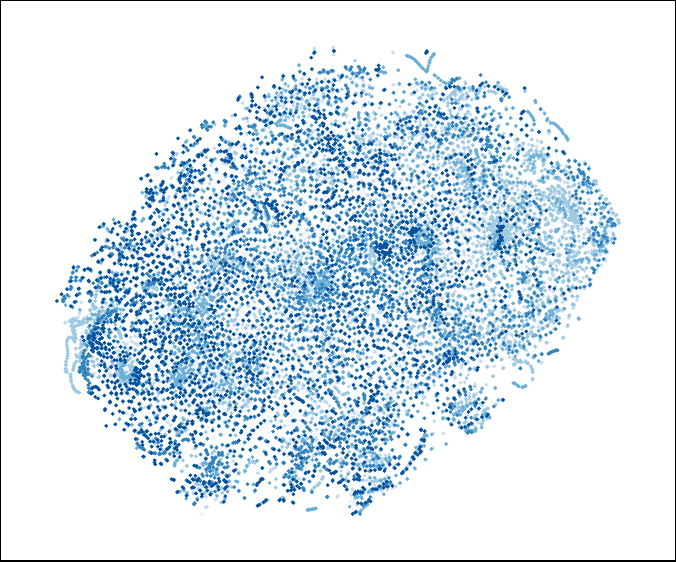}
            \caption[]%
            {{\small O$\rightarrow$I : Adversarial Transfer}}    
            \label{fig:tsne_o2i_atl}
        \end{subfigure}
        \vskip\baselineskip
        \begin{subfigure}[b]{0.475\textwidth}   
            \centering 
            \includegraphics[width=\textwidth]{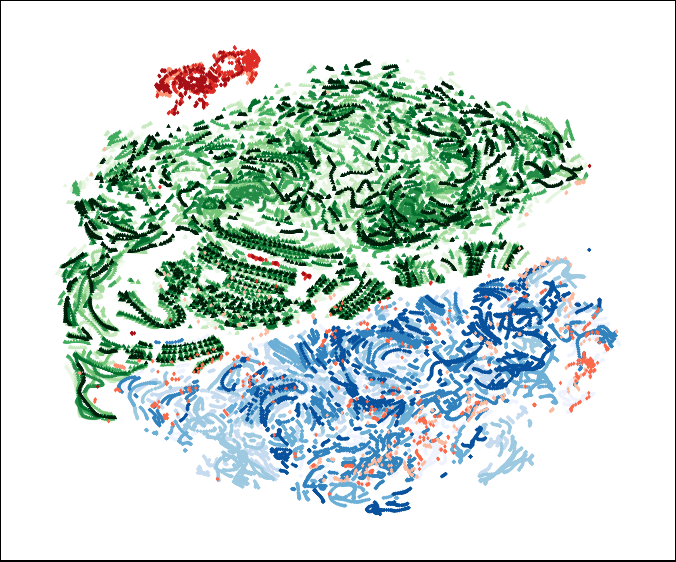}
            \caption[]%
            {{\small IOT$\rightarrow$I : Standard Transfer}}    
            \label{fig:tsne_iot2i_tl}
        \end{subfigure}
        \hfill
        \begin{subfigure}[b]{0.475\textwidth}   
            \centering 
            \includegraphics[width=\textwidth]{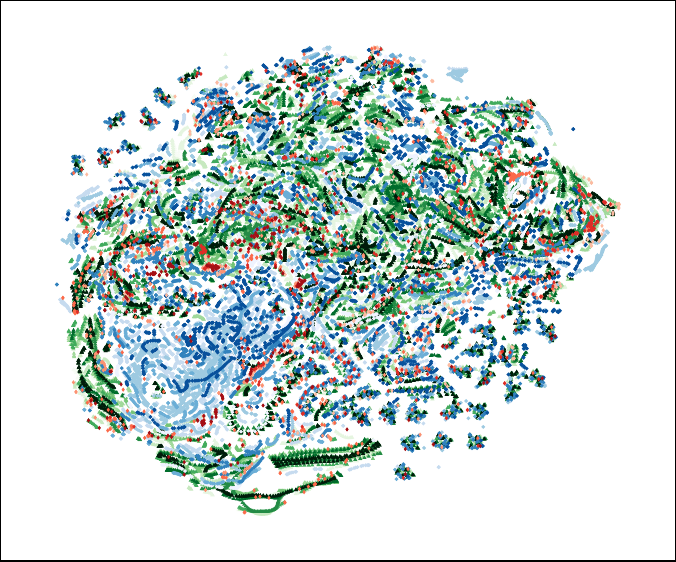}
            \caption[]%
            {{\small IOT$\rightarrow$I : Adversarial Transfer}}    
            \label{fig:tsne_iot2i_atl}
        \end{subfigure}
        \caption[ The average and standard deviation of critical parameters ]
        {\small 
        t-SNE visualization of the features for the transfer scenarios O$\rightarrow$I (top) IOT$\rightarrow$I (bottom) and for a standard (left) and adversarial transfer (right). The features represented are those of the source patients, the patient IDIAB \#1 having been arbitrarily chosen as the target patient. The two-dimensional representation of the features was obtained by t-SNE after reducing the original dimension to 50 through principal component analysis (PCA). Each color represents a patient and each shade of color represents a dataset (red for the IDIAB, blue for OhioT1DM and green for T1DMS). The colors have been chosen according to \cite{brewer1994color}.} 
        \label{fig:tsne}
\end{figure*}

\RestyleAlgo{ruled}
\LinesNumbered

\begin{algorithm}
 \KwData{features F and domain of S samples, neighbourhood size N, number of domains D}
 \KwResult{Local Domain Perplexity (LDP)}
 \For{$i=1 \rightarrow S$}{
 $F_i \leftarrow$ features of sample i\;
  $neigh_i \leftarrow$ the N samples that minimize the Euclidian distance of their features with $F_i$\;
  $P(d), d \in [1..D] \leftarrow 1/N \cdot \sum_1^N \text{count}(neigh_i = d)$\;
  $LDP_i\leftarrow  1/D \cdot 2^{\sum_d P(d) \cdot \log_2(P(d))}$\;
 }
 $LDP \leftarrow 1/S \cdot \sum_{i=1}^{S}{LDP_i}$ \;
 \KwRet{$LDP$}\;
 \caption{Computation of the LDP metric}
 \label{algo:NNP}
\end{algorithm}

\newcolumntype{Z}[1]{>{\centering}m{#1}}

\begin{table}
    \footnotesize
    \caption{Mean (with standard deviation) Local Domain Perplexity (LDP) of the features extracted by the global models for every Source (S) / Targer (T) datasets combinations.}
    \label{table:atl_perp}
    \begin{tabularx}{\linewidth}{c|c||C|C|C}
            \toprule
            
            
            \multicolumn{2}{c||}{\textbf{Dataset}} & \multirowcell{2}{\textbf{Standard}\\\textbf{Transfer}\\\textbf{(global)}} & \multirowcell{2}{\textbf{Adversarial}\\\textbf{Transfer}\\\textbf{(global)}} & \multirowcell{2}{\textbf{Relative}\\\textbf{Change}\\\textbf{(in \%)}}\\
            \cmidrule{1-2}
            \textbf{S} & \textbf{T} &  & & \\

            \midrule
            
            I & \multirow{7}{*}{I} & 0.72 (0.04) &  \underline{\textbf{0.79 (0.01)}} & +11.06 (6.58)\\
            O &  & 0.50 (0.02) & \underline{\textbf{0.68 (0.01)}} & +36.93 (5.20)\\
            T &  & 0.72 (0.01) & \underline{\textbf{0.73 (0.01)}} & +1.15 (2.09)\\
            IO &  & 0.34 (0.01) & \underline{\textbf{0.53 (0.01)}} & +54.28 (7.00)\\
            IT &  & 0.47 (0.00) & \underline{\textbf{0.51 (0.01)}}& +10.29 (2.58)\\
            OT &  & 0.38 (0.00) & \underline{\textbf{0.46 (0.02)}}& +21.53 (5.70)\\
            IOT &  & 0.29 (0.00) & \underline{\textbf{0.40 (0.01)}}& +38.99 (3.03)\\
            
            \midrule
            
            I & \multirow{7}{*}{O} & 0.67 (0.01) & \underline{\textbf{0.78 (0.01)}}& +15.48 (1.71)\\
            O &  & 0.55 (0.04) & \underline{\textbf{0.71 (0.02)}}& +29.05 (915)\\
            T &  & 0.72 (0.00) & \underline{\textbf{0.72 (0.01)}}& +0.28 (2.01)\\
            IO &  & 0.35 (0.02) & \underline{\textbf{0.52 (0.01)}}& +47.51 (7.02)\\
            IT &  & 0.43 (0.00) & \underline{\textbf{0.49 (0.01)}}& +13.84 (3.10)\\
            OT &  & 0.41 (0.00) & \underline{\textbf{0.49 (0.03)}}& +21.80 (7.86)\\
            IOT &  & 0.29 (0.01) & \underline{\textbf{0.40 (0.01)}}& +35.63 (4.92)\\
            
            \bottomrule
        \end{tabularx}
\end{table}


In this subsection, we aim at shedding some light on how the adversarial methodology manages to improve the performances of the model over a standard one. First, through Fig. \ref{fig:tsne}, we can look at the t-SNE 2-D representations of the features \cite{maaten2008visualizing} outputted by the feature extractor module (see Section \ref{fig:atl_glucose}), in the IOT$\rightarrow$I and O$\rightarrow$I transfer scenarios. Then, we propose a new metric to quantify, on the raw features, the observations we make on the t-SNE representations.

Within Figure \ref{fig:tsne_o2i_tl}, we can notice that the features of a given patient (the same shade of a given color) are often grouped in clusters. This shows that the model manages to discriminate the patients by extracting very patient-specific features. Conversely, the t-SNE representation of the features computed by the adversarial methodology displayed by Figure \ref{fig:tsne_o2i_atl} is much more diffuse indicating more general features within the source patients.

By looking at transfers with several source datasets through Figure \ref{fig:tsne_iot2i_tl} and \ref{fig:tsne_iot2i_atl}, we observe the same behavior which is, this time, more pronounced. While the three datasets occupy very delimited regions in the 2D space in the case of standard transfer, they become interlaced using adversarial transfer. This shows that the extracted features are much more general for patients within the same dataset, but also more general across datasets. By being thus more general in general, these features are therefore more easily transferable to a new unknown patient. This also validates our initial hypothesis: in the \textit{non-adversarial} multi-source transfer scenario, if it is beneficial, the model will learn to tell the domains apart and specialize its processing to the patients it identifies.


In order to quantify the observations made in the 2D space in the original feature space, we propose a new metric named Local Domain Perplexity (LDP). It measures in average, from 0 to 1, how uniform the distribution of the features' domains (here the patients) is in their close neighborhood. Whereas a high LDP implies that the features are general across all the domains, a low LDP means that the features are very specialized to the domains. It is computed following these steps, for which Algorithm \ref{algo:NNP} provides the pseudo-code:
\begin{enumerate}
    \item The distances between all pairs of feature samples are computed;
    \item For each sample, the $N$ closest samples are identified as the neighbors, $neigh_i$, $i \in [1,N],$ of this sample;
    \item  For each sample, the domain probability distribution of the neighbors is computed : $P(d) = 1/N \cdot \sum_1^N \text{count}(neigh_i = d)$;
    \item The LDP is computed as the perplexity, $2^{\sum_d P(d) \cdot \log_2(P(d))}$, rescaled between 0 and 1, and averaged over all the samples.
\end{enumerate}

Table \ref{table:atl_perp} provides the LDP measurements for each transfer scenario of the global models. The use of the adversarial methodology is shown to increase the LDP for all the transfer scenarios. The improvement are the strongest for the \textit{inter}, \textit{intra+inter}, and \textit{intra+inter+synth} scenarios and the weakest for the \textit{synth} scenario, in accordance with the accuracy improvements displayed by Figure \ref{fig:atl_vs_tl}. 
\section{Conclusion}

In this study, we proposed a multi-source adversarial training methodology for transfer learning in healthcare, field where deep learning struggles to perform due to the lack of data. We demonstrate its interest for the challenging task of glucose forecasting for diabetic people, characterized by the high inter-/intra variability of its population and the need of personalized models.

We evaluate the proposed approach by comparing it to a standard transfer methodology with three highly diverse datasets (different diabetes types, experimental settings, real and virtual patients). First, we show that transfer learning is an effective way to solve the lack of data for the training of deep models in the field of glucose prediction, even if the source patients are of a different diabetes type or have their data collected in different experimental settings. Then, we demonstrate the statistical and clinically significant superiority of our adversarial transfer methodology. While the transfer is mostly successful when the source data are closely related to the target data (e.g., same datasets), it can be improved by augmenting the source data with other, less related data (similar disease, different experimental settings, or even synthetic data).

Furthermore, we analyzed the behavior of the adversarial training methodology in the learning of the model's feature representation. We show visually and empirically that it improves the generality of the extracted features across the source patients, from which the improved performances, after finetuning to the target patient, are derived. To this end, we have introduced a new metric, called Local Domain Perplexity (LDP), that measures how uniform the domain distribution is in the locality of the samples' features.

Looking forward, given the slight disparity between the results obtained on the IDIAB and the OhioT1DM dataset, we think that the results could be improved by working on the meticulous selection and balancing of the source data. Also, we could improve the use of synthetic data as the source (or part of) of the transfer by designing specific simulation scenarios in this regard.

We hope that these findings will open new ways of tackling challenges related to the forecasting of glucose for diabetic people, and those related to healthcare in general, which is characterized by its highly heterogenous data.

\section*{Acknowledgment}

This  work  is  supported  by  the  "IDI  2017"  project  funded by the IDEX Paris-Saclay, ANR-11-IDEX-0003-02. We would like to thank the French diabetes health network Revesdiab and Dr. Sylvie JOANNIDIS for their help in building the IDIAB dataset used in this study.

\bibliography{bibtex.bib}
\bibliographystyle{IEEEtran}

\ifCLASSOPTIONcaptionsoff
  \newpage
\fi

\end{document}